\documentclass[runningheads]{llncs}
\usepackage{graphicx}
\usepackage{amsmath}
\usepackage{amssymb}
\usepackage{amsfonts}
\usepackage{multirow}
\usepackage{url}
\usepackage{booktabs}
\usepackage[ruled,vlined, linesnumbered]{algorithm2e}
\usepackage{wrapfig}
\usepackage{lipsum}

\include{pythonlisting}

\newcommand{\approach}[1]{\emph{#1}}

\newcommand{\etal}{\textit{et al}. }

\newcommand\blfootnote[1]{%
  \begingroup
  \renewcommand\thefootnote{}\footnote{#1}%
  \addtocounter{footnote}{-1}%
  \endgroup
}

\usepackage{enumitem}
\setitemize{noitemsep,topsep=1pt,parsep=1pt,partopsep=1pt}
\makeatletter
\def\BState{\State\hskip-\ALG@thistlm}
\makeatother
\begin{document}
\title{Brain Segmentation from $k$-space with End-to-end Recurrent Attention Network}

\author{Qiaoying Huang\inst{1*} \and
Xiao Chen\inst{2}\and
Dimitris Metaxas\inst{1}\and
Mariappan S. Nadar\inst{2}}


\institute{
Rutgers University, Department of Computer Science, Piscataway, NJ, USA \and
Siemens Healthineers, Digital Technology and Innovation, Princeton, NJ, USA\\
}

\maketitle              
\begin{abstract}
The task of medical image segmentation commonly involves an image reconstruction step to convert acquired raw data to images before any analysis. 
However, noises, artifacts and loss of information due to the reconstruction process is almost inevitable, which compromises the final performance of segmentation. 
We present a novel learning framework that performs magnetic resonance brain image segmentation directly from $k$-space data. 
The end-to-end framework consists of a unique task-driven attention module that recurrently utilizes intermediate segmentation estimation to facilitate image-domain feature extraction from the raw data, thus closely bridging the reconstruction and the segmentation tasks.
In addition, to address the challenge of manual labeling, we introduce a novel workflow to generate labeled training data for segmentation by exploiting imaging modality simulators and digital phantoms. 
Extensive experimental results show that the proposed method outperforms several state-of-the-art methods.\blfootnote{*Work done while intern at Siemens Healthineers.}
\end{abstract}

\section{Introduction}\label{sec:intro}
Most image segmentation tasks start from existing images. While this might seem self-evident for natural image applications, many medical imaging modalities do not acquire data in the image space.
Magnetic Resonance Imaging (MRI), for example, acquires data in the spatial-frequency domain (the so called $k$-space) and the MR images need to be reconstructed from the $k$-space data before further analysis. 
The traditional pipeline of image segmentation treats reconstruction and segmentation as separate tasks. 
Image noises, residual artifacts and potential loss of information on the imperfect reconstructed images are almost inevitable, even with advanced image reconstruction methods. 
On the other hand, these algorithms are usually designed to recover images for optimal visual quality to be used by physicians, rather than the ``task'' quality, namely the ``segmentation quality'' here. 
Without the final segmentation quality as a target, the reconstruction algorithm may discard image features that are critical for segmentation but less influential to image quality. 
Meanwhile, the algorithm may spend most of the resources (e.g. reconstruction time) to recover image features that are less important to segmentation accuracy improvement. 
It is thus highly desirable to use an end-to-end approach to predict segmentation directly from $k$-space. 

\begin{figure*}[!t]
    \centering    
    \includegraphics[width=\textwidth]{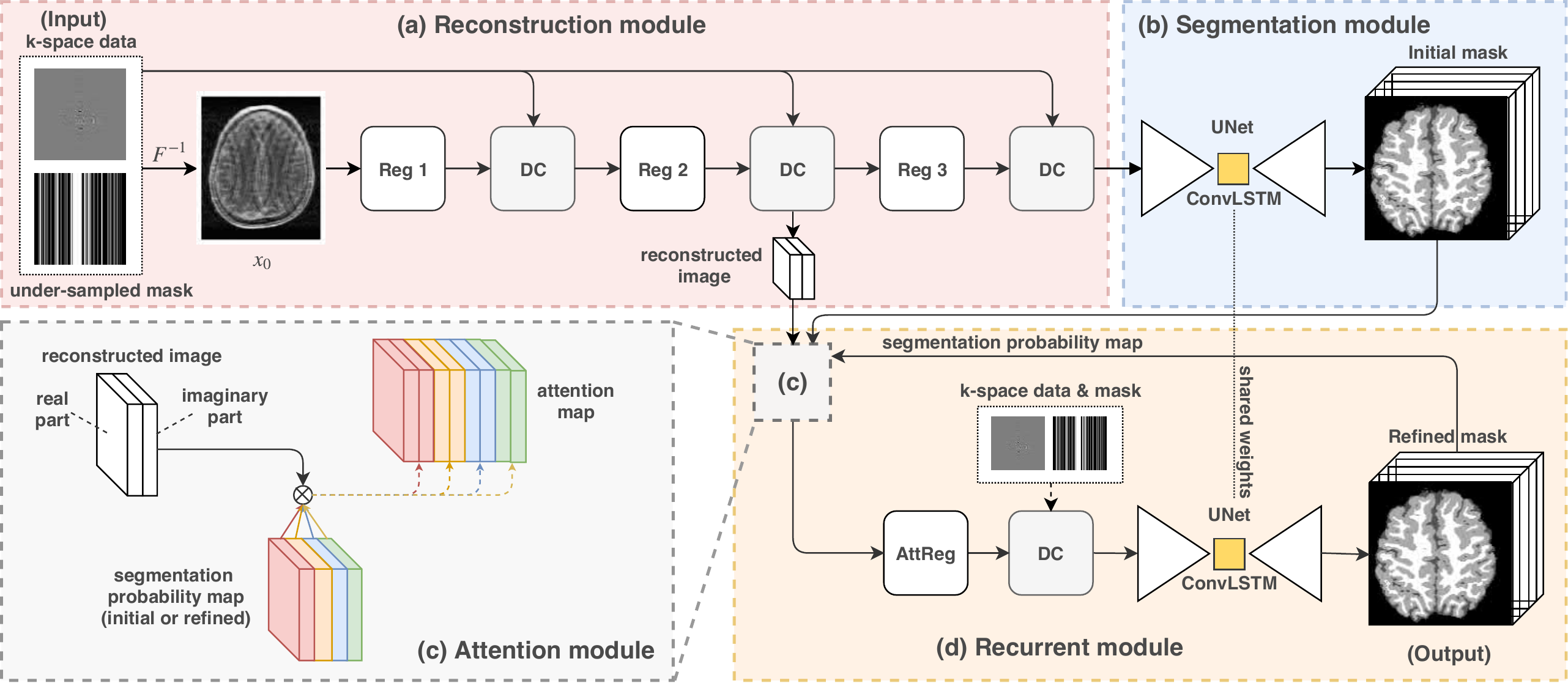}\\
    \vspace{-0.15in}
    \caption{
    The proposed model takes the under-sampled k-space data as input and outputs segmentation masks. 
    It consists of (a) reconstruction module, (b) segmentation module, (c) attention module and (d) recurrent module.
    }
    \label{fig:network}
\end{figure*}

Several end-to-end learning frameworks have been proposed for various applications. 
Caballero \etal \cite{caballero2014application} propose an unsupervised brain segmentation method that treats both reconstruction and segmentation simultaneously using patch-based dictionary sparsity and a Gaussian mixture model. 
This method is not  suitable for complicated scenarios such as the case  where there are no clear boundaries between different tissues.
Schlemper \etal \cite{schlemper2018cardiac} propose two neural networks (LI-net and Syn-net) that predict cardiac segmentation from under-sampled $k$-space data.
The Syn-net uses UNet \cite{ronneberger2015u} to map zero-filling image (inverse Fourier transform of under-sampled $k$-space) directly to segmentation maps. 
The LI-net exploits latent space features and requires fully-sampled images during training. 
Huang \etal \cite{huang2019fr} propose a Joint-FR-Net that simply optimizes a joint reconstruction and segmentation network by a combined loss function.
The most relevant work to ours is \cite{jointcsliyan2018} where the authors present a deep neural network architecture termed SegNetMRI. Specifically, the reconstruction and the segmentation sub-networks are pretrained and fine-tuned with shared encoders. The final segmentation depends on the combination of the intermediate segmentation results.
The segmentation sub-network in SegNetMRI is trained on reconstructed images, which contain artifacts and noises and may influence the performance of segmentation.
Different from SegNetMRI, we propose a new feature sharing method that overcomes the interference with noisy data.

One challenge for end-to-end segmentation learning is the preparation of the training data with ground truth segmentation. 
Most current studies simulate the raw $k$-space data from DICOM images using direct Fourier transform. 
However, realistic $k$-space data can rarely be recovered from the images alone due to the complex-value nature of MR and common MR post-processing practices that may alter the acquisition.    
In addition, it is hard to obtain the ground truth segmentation for training. 
Manual labeling which is performed on images is prone to imperfect reconstruction and human error for small anatomy structures. 

We present here an end-to-end architecture: Segmentation with End-to-end Recurrent Attention Network (\approach{SERANet}) featuring a unique recurrent attention module that closely connects the reconstruction and the segmentation tasks. 
The intermediate segmentation is recurrently exploited as anatomical prior that provides guidance to recover \emph{segmentation-driven} image features from the raw data, which in turn improves segmentation performance.
Our contributions include three folds: a) We propose an approach that recurrently performs image segmentation directly from under-sampled $k$-space data;
b) We introduce a novel attention module that guides the network to generate segmentation-driven image features to improve the segmentation performance;
c) We present a novel workflow to generate under-sampled $k$-space data with oracle segmentation maps by exploiting an MRI simulator and digital brain phantoms. 

\section{Methods}\label{sec:method}
In this section, we begin with a brief introduction to the problem of achieving segmentation directly from $k$-space. 
Then we describe our attention module and end-to-end recurrent framework.
\subsection{Background}
Segmentation from raw data can be generally divided into two subproblems: reconstruction and segmentation.
For deep learning based reconstruction on under-sampled $k$-space: $x=f_{Rec}(y,m)$, where $x\in \mathbb{R}^{2\times w\times h}$ is the reconstructed image (real and imaginary parts concatenated in the first dimension), $y$ is the under-sampled $k$-space data and $m$ is the under-sampling mask indicating the position of sampling. 
$f_{Rec}$ is a deep neural network that can be optimized by reconstruction loss, such as the $l_2$ loss: $l_2(x, x_{gt})=\|x-x_{gt}\|_2$.
Specifically, as shown in Figure \ref{fig:network} (a), the reconstruction module consists of two basic components. 
One component is a Data Consistency (\approach{DC} layer) \cite{schlemper2017deep} that compensates the difference between the estimated and the measured $k$-space data. 
The other component is a regularization block (\approach{Reg} block) that takes as input the zero-filled fast Fourier transform reconstructed image  $x_0$, or the output from the \approach{DC} layer $x_{dc}$, and outputs an image $x$.
Cascaded CNN \cite{schlemper2017deep} and UNet \cite{huang2018mri,jointcsliyan2018} are two popular choices for \approach{Reg} block. 
Deep learning based iterative reconstruction, motivated by the recent success of compressed sensing in MR image reconstruction, is realized by cascading a series of \approach{Reg} blocks and \approach{DC} layers.
The second step is to use the reconstructed image $x$ to predict segmentation probabilities: $s=f_{Seg}(x)$.
$f_{Seg}$ is also a deep neural network that can be optimized by segmentation loss, such as the cross entropy loss: $l_{ce}(s, s_{gt}) = -\sum_{s^i\in S}s_{gt}^i\log(s^i)$. 
As shown in Figure \ref{fig:network} (b), the segmentation module is usually a UNet shape \cite{ronneberger2015u}.
\subsection{Attention module}
In end-to-end training, it is beneficial to share information among different tasks.
Therefore, we propose an attention module, as shown in Figure \ref{fig:network} (c) that bridges the gap between reconstruction and segmentation to facilitate learning segmentation-aware features in the image domain. 
We consider a brain segmentation map as anatomical prior and use it to guide the image reconstruction such that the segmentation information is explicitly utilized to extract segmentation-aware image features from the raw $k$-space data. 
We use an attention network to facilitate the segmentation-aware learning. 
Different from the traditional attention mechanism that only considers two classes in one forward pass \cite{li2018tell}, we propose to generate multi-class attention maps simultaneously to distinguish features among four classes in the human brain: cerebrospinal fluid (CSF), gray matter (GM), white matter (WM) and background. 
After one forward pass through the image reconstruction module and the segmentation module, an initial segmentation result is obtained (see Figure \ref{fig:network} part (b)). The segmentation maps $s\in \mathbb{R}^{4\times w \times h}$ have four tissue maps in separate channels, which are concatenation along the first dimension: 
$s = s^1\oplus  s^2\oplus  s^3\oplus  s^4$,
where $s^i$ indicates the $i^{th}$ class prediction, $\oplus$ represents concatenating along the first dimension. 
The segmentation map itself is already a probability map. After a softmax layer $\sigma(\cdot)$ that ensures the sum of the four different classes to be $1$, the maps can be utilized directly for attention. 
Each of the four segmentation probability maps are element-wise multiplied with the input image features $x_{t-1}\in \mathbb{R}^{2\times w\times h}$ to generate new features $x_{t}\in \mathbb{R}^{8\times w\times h}$:
\begin{equation}\label{equ:x}
    x_t = (s_{t-1}^1 \odot x_{t-1}) \oplus (s_{t-1}^2 \odot x_{t-1}) \oplus ( s_{t-1}^3 \odot x_{t-1}) \oplus  (s_{t-1}^4 \odot x_{t-1}),
\end{equation}
where subscript $t$ represents the $t^{th}$ intermediate result (explained in the next section).
As shown in Figure \ref{fig:network} (c), the new image features $x_t$ go through one \approach{Reg} block for attention features (referred as \approach{AttReg}) and one \approach{DC} layer, in order to extract image features. 
The difference between \approach{AttReg} and \approach{Reg} in the reconstruction module is the input channel size: \approach{AttReg} has $8$ instead of $2$ channels. 
The output of the attention-assisted image feature extraction is then fed to the same segmentation module with shared weights to generate a new segmentation estimation $s_{t}$. Formally, $s_{t}$ can be expressed as follows. 
\begin{equation}\label{equ:news}
    s_{t} = (f_{Seg}\circ f_{DC}\circ f_{AttReg})(x_t),
\end{equation}
where $\circ$ denotes function composition and $s_{t}$ is the new segmentation estimation. 
By explicitly utilizing intermediate segmentation results for reconstruction, or more precisely image feature extraction, segmentation-driven features will be generated, which in turn improves segmentation performance during training with back-propagation algorithm. 
It can be seen from Figure \ref{fig:exp_fig} that clear boundaries are generated using the proposed \approach{SERANet} from under-sampled $k$-space data, while the ground truth reconstruction from fully-sampled $k$-space data contains noise. 

\subsection{Recurrent framework}
We treat segmentation feature learning as a recurrent procedures that final result is achieved by iterating the attention module several times.
Formally, given under-sampled $k$-space data $y$ and mask $m$, \approach{SERANet} learns to segment the brain in $T$ iterations.
$N$ is the number of \approach{Reg} blocks and \approach{DC} layers. 
As described in Algorithm 1, line 1 and 2 generate the initial reconstructed image $x_0$ and brain tissue map $s_0$.
Then line 3 to 5 represent a recurrent segmentation-aware reconstruction and segmentation process.
The attention module $f_{AttReg}$ takes the initial reconstruction feature $x_0^{N-1}$ (feature from the $N-1$ reconstruction block) and segmentation probability maps $s_{t-1}$ as input and generates new image $x_{t}$, as illustrated in Figure \ref{fig:network} (c) and (d).
To capture and memorize the spatial information at different recurrences, a ConvLSTM layer \cite{xingjian2015convolutional} is integrated into the UNet for segmentation.
The objective function of the whole model is defined as ${l}_{ce}(s_T, s_{gt})$, where $s_T$ denotes the output of the final iteration.
By doing so, the reconstruction module in our method does not see nor need any ground truth reconstruction image during training.
The recovered image content is guided by the segmentation error solely, which suffices the aim to recover image domain features from the raw data that best suits the segmentation task, rather than the conventional reconstruction task. 
The usage of ``reconstruction'' to name the module is just for conceptual simplicity.

\begin{algorithm}[t!] 
\SetKwData{Left}{left}\SetKwData{This}{this}\SetKwData{Up}{up}
\SetKwFunction{Union}{Union}\SetKwFunction{FindCompress}{FindCompress}
\SetKwInOut{Input}{input}\SetKwInOut{Output}{output}
        \SetCustomAlgoRuledWidth{0.45\textwidth}  
        \caption{\approach{SERANet}: Segmentation with Recurrent Attention Network}
          \Input{under-sampled $k$-space data $y$, under-sampling mask $m$, $N$, $T$}
          $x_{0}^{(N-1)}, x_{0}^{(N)}\leftarrow f_{Rec}(y,m)$ \tcp*[r]{initial reconstruction feature $x_0$}
          $ s_{0}\leftarrow f_{Seg}(x_{0}^{(N)})$ \tcp*[r]{initial segmentation result $s_0$}
          \If{$t \leq T$}{ 
          $x_{t} \leftarrow  f_{DC}\circ f_{AttReg}(x_{0}^{(N-1)}, s_{t-1})$ \tcp*[r]{Attention module}
          $s_{t} \leftarrow f_{Seg}(x_{t})$ \tcp*[r]{Recurrent segmentation}
        }
        \Output{$s_T$}
\label{alg:seg}
\end{algorithm}

\subsection{Generate $k$-space data with oracle segmentation maps}
We propose here a novel method to generate realistic $k$-space data with ground truth segmentation map.  
Specifically, a widely utilized MRI scanner simulator MRiLab \cite{liu2017fast}\cite{mrilab} is adopted to provide a realistic virtual MR scanning environment, which includes scanner system imperfection, MR acquisition pattern and MR data generation and recording.
We use publicly available digital brain phantoms from BrainWeb \cite{chris1997Brainweb} as the object ``scanned'' in the MRI simulator. Each brain is consisted of 12 tissue types with known spatial distributions. Each tissue type has a unique set of values of MR physical parameters such as T1 and T2 that are needed for the MR scan simulation. Fully-sampled $k$-space data is then simulated by scanning the digital brain in MRiLab. Under-sampling is performed retrospectively by keeping a subset of the full-sampled data. To mimic realistic MR scanning, white Gaussian noises are added to the $k$-space data at multiple levels. The network never sees the fully-sampled data. The spatial distributions of the tissues are the oracle segmentation maps. 

\section{Experiments}
\begin{figure*}[!t]
    \centering    
    \includegraphics[width=0.49\textwidth]{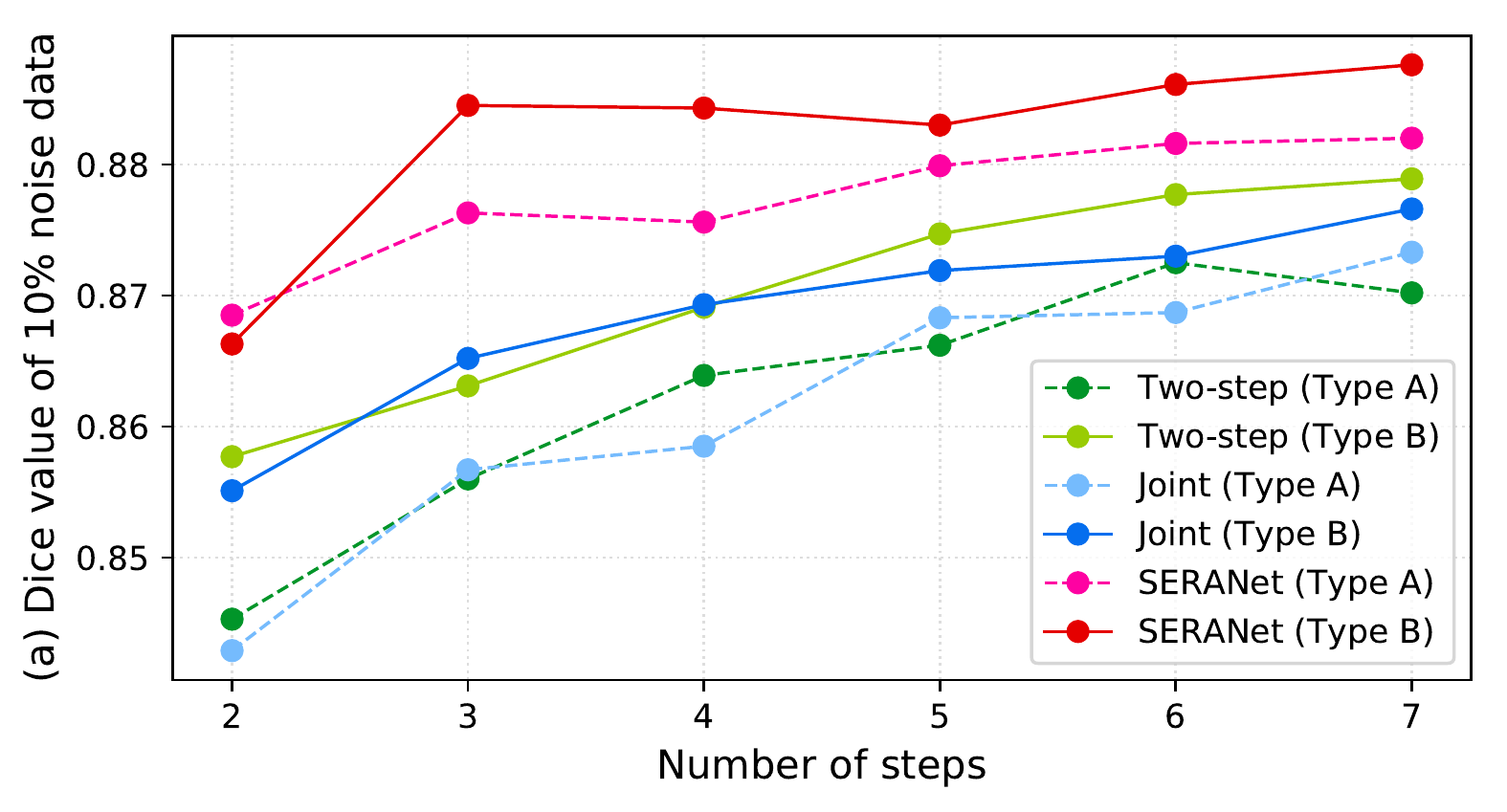}
    \includegraphics[width=0.49\textwidth]{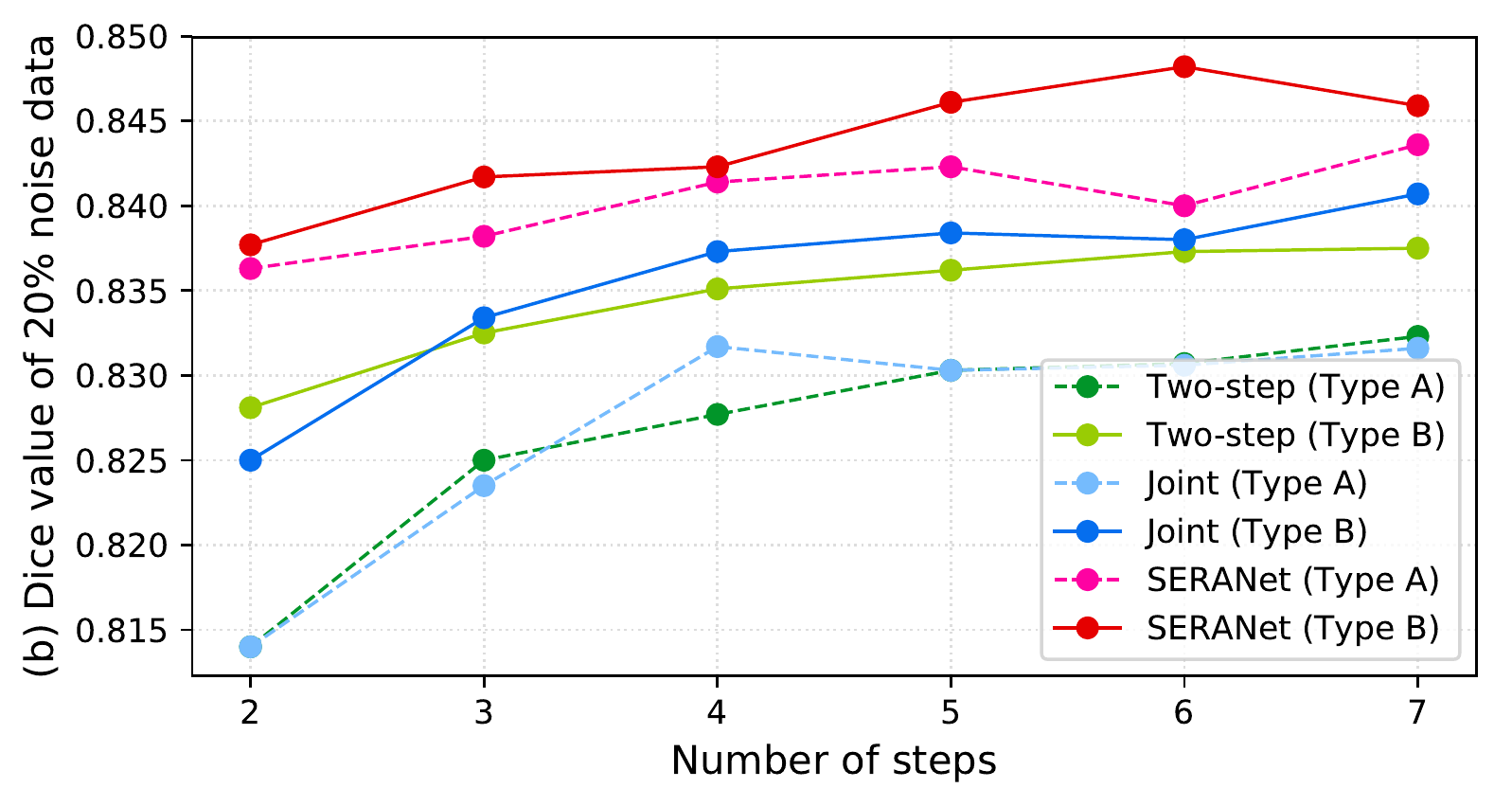}\\
    \vspace{-0.1in}
    \caption{Dice score as a function of the number of reconstruction blocks.}
    \label{fig:exp_step}
    \vspace{-0.1in}
\end{figure*}

\paragraph{\textbf{Implementation Details}}
Total 20 healthy digital 3D brain volumes are utilized. The networks are trained on 2D axial slices. $969$ slices of $17$ brains are used for training and the rest $171$ slices from $3$ brains are reserved for testing only.
Each digital brain is scanned by a spin echo sequence with Cartesian readout. Average TE $=80$ ms and TR $=3$ s are used and $5\%$ variation of both TE and TR values are introduced for varying MR contrasts. 
Each slice has the corresponding tissue segmentation mask from BrainWeb.
All slices have a unified size of $180\times216$ with 1 mm isotropic resolution. 
We use a zero-mean Gaussian distribution with a densely sampled $k$-space center to realize a pseudo-random under-sampling pattern, where $30\%$ phase encoding lines are maintained with 16 center $k$-space lines. All k-space data are added with additional 10\% and 20\% white Gaussian noise.
All models are implemented in Pytorch and trained on NVIDIA TITAN Xp.
Hyperparameters are set as: a learning rate of $10^{-4}$ with decreasing rate of $0.5$ for every 20 epochs, 50 maximum epochs, batch size of 12.
Adam optimizer is used in training all the networks. 
We adopt Dice's score as evaluation metrice in all experiments.
For the recurrent steps $T$, the segmentation performance of our model in terms of Dice's score has converged after two recurrences. So we empirically set $T=2$. 
\begin{table}[!t]
\center
\setlength{\tabcolsep}{2pt}
\renewcommand{\arraystretch}{1.4}
\scriptsize
\caption{Dice's score  of \approach{SERANet} trained with different losses}
\label{tab:loss}
\begin{tabular}{c|cccl|cccl}
\toprule
\multirow{2}{*}{Loss} & \multicolumn{4}{c}{10\% noise} & \multicolumn{4}{|c}{20\% noise} \\
                         & CSF   & GM   & WM   & Aver.  & CSF   & GM   & WM   & Aver.  \\
\hline
$l_{ce}(s_T)+l_2(x_T)$     & 0.8048 & 0.8841 & 0.8518 & 0.8469 & 0.7995  & 0.8751 & 0.8092 & 0.8279  \\
$\sum_{t=0}^{T}l_{ce}(s_t)$      & \textbf{0.8513} & 0.9082  & 0.8796 & 0.8797 &    0.8041   & 0.8733 & 0.8283 & 0.8352  \\
$l_{ce}(s_T)$      & 0.8482 & \textbf{0.9102} & \textbf{0.8814} &  \textbf{0.8799} & \textbf{0.8083} & \textbf{0.8762} & \textbf{0.8415} & \textbf{0.8423}\\
\bottomrule
\end{tabular}
\vspace{-0.1in}
\end{table}
\paragraph{\textbf{Effect of different losses}}
We provide a comparison of training \approach{SERANet} with different losses in Tabel \ref{tab:loss}.
We observe that \approach{SERANet} constrained by the reconstruction loss ($l_2$) performs the worst, as shown in the first row of Table \ref{tab:loss}.
This may seem surprising at first but it actually verifies that reconstruction from the images with noise and artifacts may compromise segmentation results. 
The model that is optimized solely using segmentation loss ($l_{ce}$) on the final segmentation estimation $s_T$ achieves the best result.
These results demonstrate two key advantages of our proposed method. 
First, due to the efficiency of the attention module, \approach{SERANet} automatically learns image feature that benefits the segmentation performance, without constraints on the reconstructed image.
Second, our method does not require cumbersome tweaking of loss weights between reconstruction and segmentation tasks.
\paragraph{\textbf{Effect of attention module}}
We design two different baselines without the attention module: one is a \approach{Two-step} model that contains separate reconstruction and segmentation modules, which are trained separately with reconstruction and segmentation losses. 
The other is a 
\approach{Joint} model, which also contains these two modules but is trained  together with only segmentation loss on the final output.
In order to evaluate the robustness under different settings, we compare performances of \approach{Two-step}, \approach{Joint} and \approach{SERANet} with different number of reconstruction blocks. 
We also implement two different reconstruction blocks: one using cascading CNN (\approach{Type A}) and the other using auto-encoder (\approach{Type B}).
The Dice's score against number of reconstruction block is plotted in Figure \ref{fig:exp_step}.
We observe, as expected, that with more cascading reconstruction blocks the segmentation performances of all tested methods improve. 
However, \approach{SERANet} outperforms the others in all settings which proves the benefit of using the attention module. 

\begin{figure*}[!t]
    \centering    
    \includegraphics[width=\textwidth]{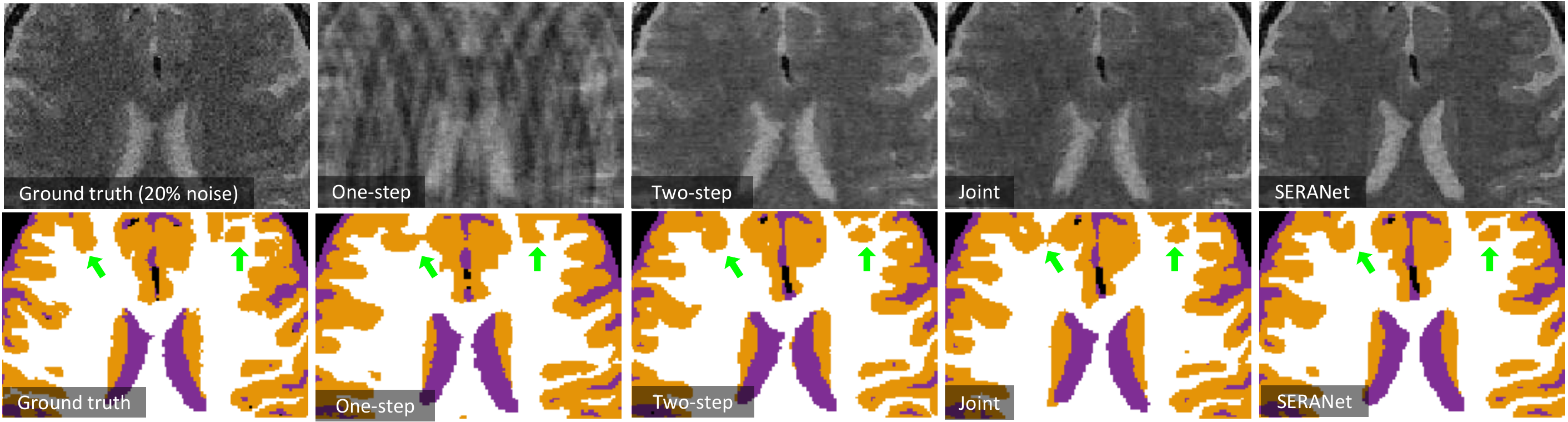}\\
    \vspace{-0.1in}
    \caption{Segmentation result of \approach{SERANet} and the compared models. CSF, GM and WM parts in each brain are colorized with purple, orange and white, respectively.}
    \label{fig:exp_fig}
    \vspace{-0.1in}
\end{figure*}

For visualization, we also train an \approach{One-step} model that takes input as zero-filling images (inverse Fourier transform of under-sampled $k$-space data) and outputs the segmentation maps. 
We visualize segmentation results and reconstructed images of models trained by 20\% noise data in Figure \ref{fig:exp_fig}.
Our method \approach{SERANet} predicts more accurate anatomical segmentation details and clearer image contrasts compared to \approach{Two-step} and \approach{Joint}. 
This shows \approach{SERANet} overcomes the interference with noisy input by using the attention module.

\begin{figure*}[!t]
    \centering    
    \includegraphics[width=\textwidth]{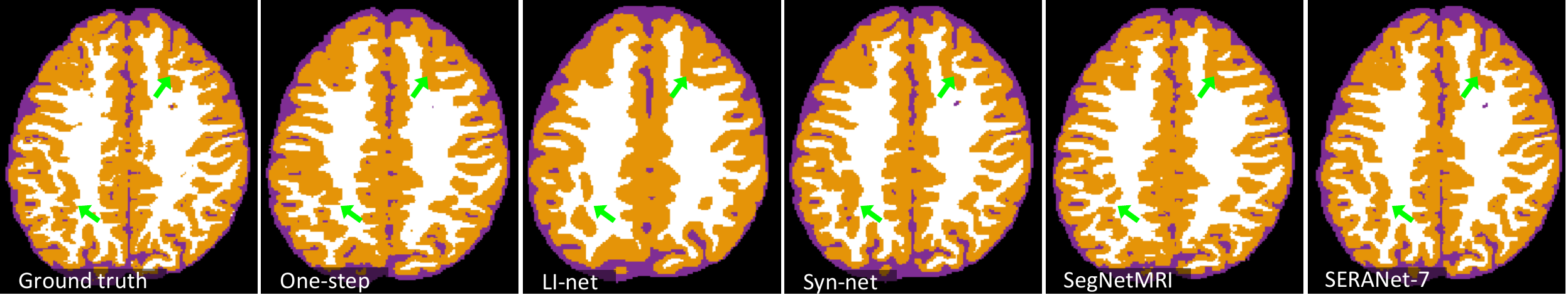}\\
    \caption{Segmentation result with different approaches.}
    \label{fig:exp_SOTA}
    \vspace{-0.2in}
\end{figure*}

\paragraph{\textbf{Comparisons to State-of-the-Art}}
We provide qualitative and quantitative comparisons to three state-of-the-art algorithms: \approach{LI-net}\cite{schlemper2018cardiac} \approach{Syn-net}\cite{schlemper2018cardiac}, and \approach{SegNetMRI}\cite{jointcsliyan2018}. We also compare to the \approach{One-step} model. 
We list the performance of \approach{SERANet-7} that with 7 reconstruction blocks and \approach{SERANet-2} that with 2 reconstruction blocks.
The results of all methods are reported in Table \ref{tab:SOTA}.
We also list whether the method is pretrained and what loss the method uses to optimize in column 2 and 3, respectively.
For \approach{LI-net} and \approach{Syn-net}, since they perform segmentation from fully-sampled data as a warm start, we consider this as a pretraining technique.
We observe that \approach{SERANet-2} and \approach{SERANet-7} consistently outperform the three state-of-the-art approaches for both 10\% and 20\% noises. 
Additionally, the Dice's scores drop more for the \approach{SegNetMRI} when noise level increases compared to \approach{SERANet}, which may be due to the fact that \approach{SegNetMRI} contains information from the noisy ground truth images. 
Example segmentation results are shown in Figure \ref{fig:exp_SOTA}. Improvements of \approach{SERANet-7} on detailed anatomy structure are highlighted by the green arrows.

\section{Conclusion}
In this paper, we propose a novel end-to-end approach \approach{SERANet} for MR brain segmentation, which performs segmentation directly on under-sampled $k$-space data via a segmentation-aware attention mechanism. 
Moreover, we design a training data generation workflow to simulate realistic MR scans on digital brain phantoms with ground truth segmentation maps. 
Extensive experiments are conducted and the results demonstrate the effectiveness and the superior performance of our model compared to the state-of-the-art methods.

\begin{table}[!t]
\center
\setlength{\tabcolsep}{1.5pt}
\renewcommand{\arraystretch}{1.4}
\scriptsize
\caption{Comparisons to State-of-the-Art}
\label{tab:SOTA}
\begin{tabular}{c|c|c|cccc|cccc}
\toprule
\multirow{2}{*}{Method} & \multirow{2}{*}{Pretrain} & \multirow{2}{*}{Loss} & \multicolumn{4}{c}{10\% noise} & \multicolumn{4}{|c}{20\% noise} \\
                        &                             &                              & CSF    & WM   & GM   & Aver.   & CSF    & WM   & GM   & Aver.   \\
                        \hline
One-step            &    No &   $l_{ce}$   &  0.7677  & 0.8334 & 0.7900  & 0.7970  & 0.7600 & 0.8324  & 0.7911  & 0.7945 \\
LI-net \cite{schlemper2018cardiac}                  &   Yes  &   $l_{ce}$  & 0.6849 & 0.7576& 0.7558  &0.7328   &  0.6686     & 0.7276     &0.7282      &0.7081  \\
Syn-net \cite{schlemper2018cardiac}                 &  Yes  & $l_{ce}$+$l_2$  &  0.7558 & 0.8256  & 0.7961 & 0.7925 & 0.7307 &0.8095 & 0.7808 & 0.7737    \\
SegNetMRI \cite{jointcsliyan2018}  &  Yes & $l_{ce}$+$l_2$  &  0.8210  & 0.8905  &0.8575   & 0.8563 & 0.7817 & 0.8472 & 0.7728  & 0.8006 \\
SERANet-2               & No & $l_{ce}$  & 0.8344 & 0.8977 & 0.8669 & 0.8663 & 0.8053 & 0.8706 & 0.8373 & 0.8377   \\
SERANet-7               & No  & $l_{ce}$  & \textbf{0.8548} & \textbf{0.9175} & \textbf{0.8905} & \textbf{0.8876} & \textbf{0.8122} & \textbf{0.8798} & \textbf{0.8457} & \textbf{0.8459} \\
\bottomrule
\end{tabular}
\end{table}

\section{Additional Implementation Details}
\begin{figure}[!h]
    \centering
    \includegraphics[width=0.25\textwidth]{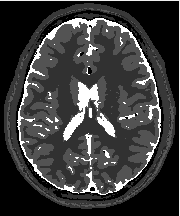}
    \hspace{0.5in}
    \includegraphics[width=0.25\textwidth]{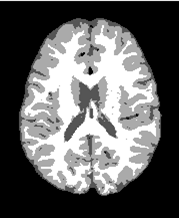}\\
    (a) \hspace{1.4in} (b) 
    \caption{(a) Original 11 tissues segment- masks. (b) Selected 4 tissues segment- masks.}
    \label{fig:segmask}
\end{figure}

In this paper, we consider a segmentation problem with four brain segmentation masks: 0-Background, 1-CSF, 2-Gray Matter and 3-White Matter.
As mentioned in the paper, these four masks are adapted from original 11 brain tissues.
The reason to use the four masks is that CSF, Gray Matter and White Matter cover most parts of the brain, as shown in Figure \ref{fig:segmask} (a) and (b).
The other eight tissues, such as vessels, skulls and skins, are grouped as Background mask.

\subsection{Data generation}






\begin{figure}[!h]
    \centering
    \includegraphics[width=0.8\textwidth]{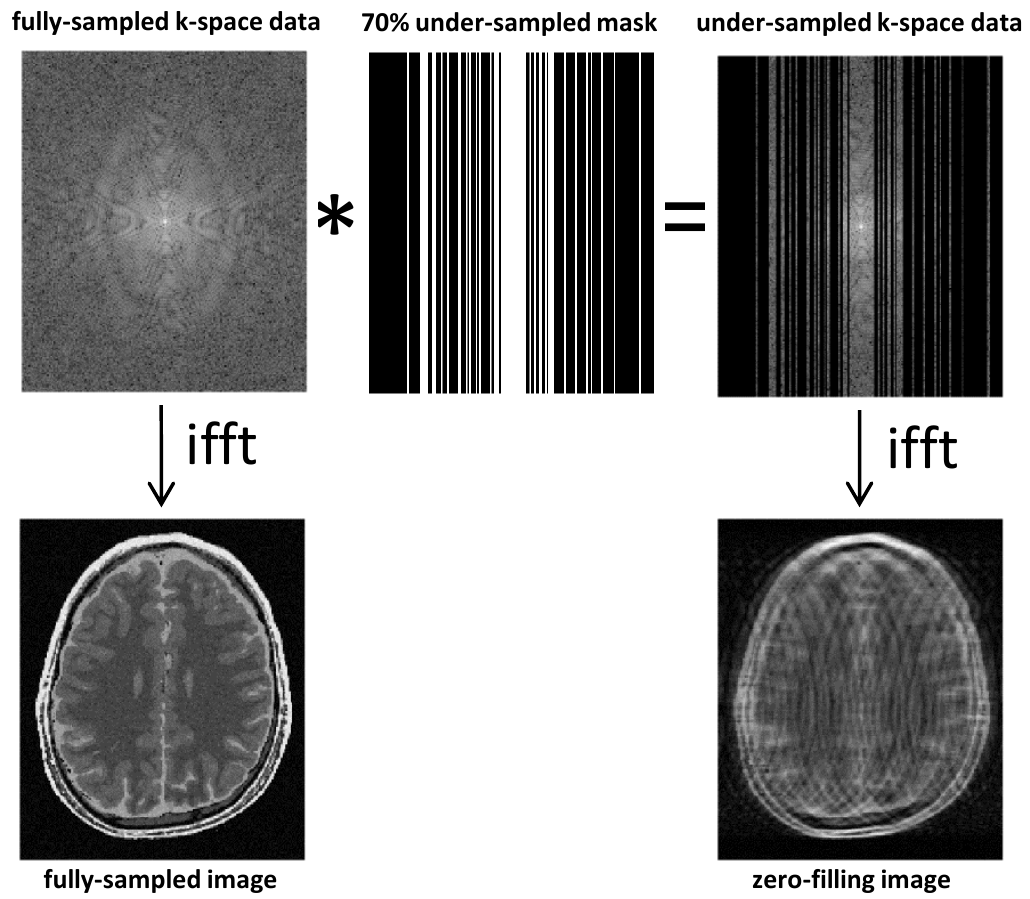}
    \caption{The process to generate under-sampled k-space data.}
    \label{fig:generation}
\end{figure}

The detailed data generation process is illustrated in Figure \ref{fig:generation}.
Given fully-sampled k-space data (top-left), we first randomly generate under-sampled mask, e.g. with 70\% sampling rate (top-middle).
Under-sampled k-space data (top-right) is obtained by employing the mask on the fully-sampled k-space data.
Then, fully-sampled image (bottom-left) can be generated from fully-sampled k-space data via inverse fast Fourier transform.
Note that the fully-sampled image is used as ground truth by some existing algorithms, however, it may contain noise and comprise the segmentation performance.
Similarly, zero-filling image (bottom-right) is generated from under-sampled k-space data via inverse Fourier transform, and is taken as the input in all models.

\subsection{Network architectures}
\begin{figure*}[!h]
    \centering
    \includegraphics[width=0.4\textwidth]{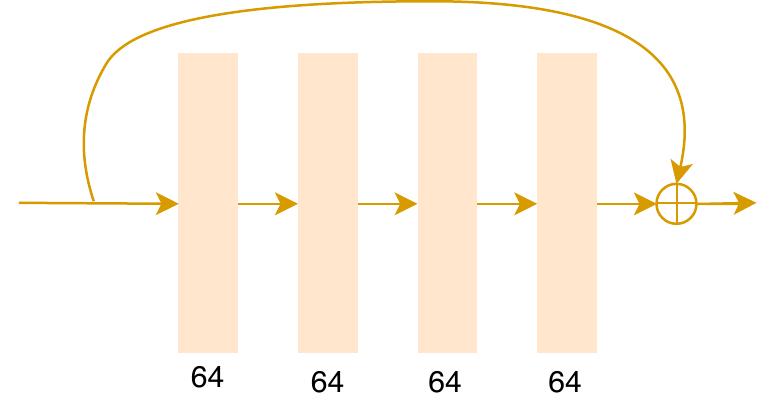}
    \hspace{0.2in}
    \includegraphics[width=0.5\textwidth]{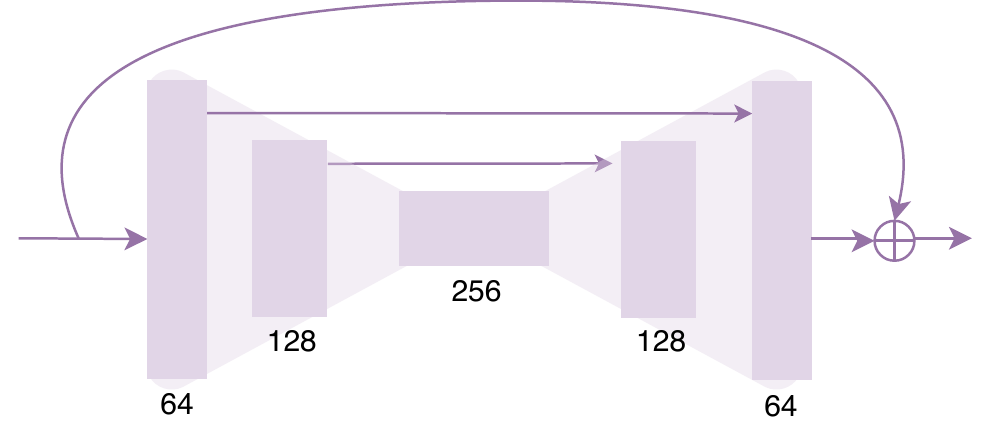} \\
    \hspace{-0.4in} (a) Regularization block: Type A \hspace{1.1in} (b) Regularization block: Type B \\
    \vspace{0.1in}
    \includegraphics[width=0.95\textwidth]{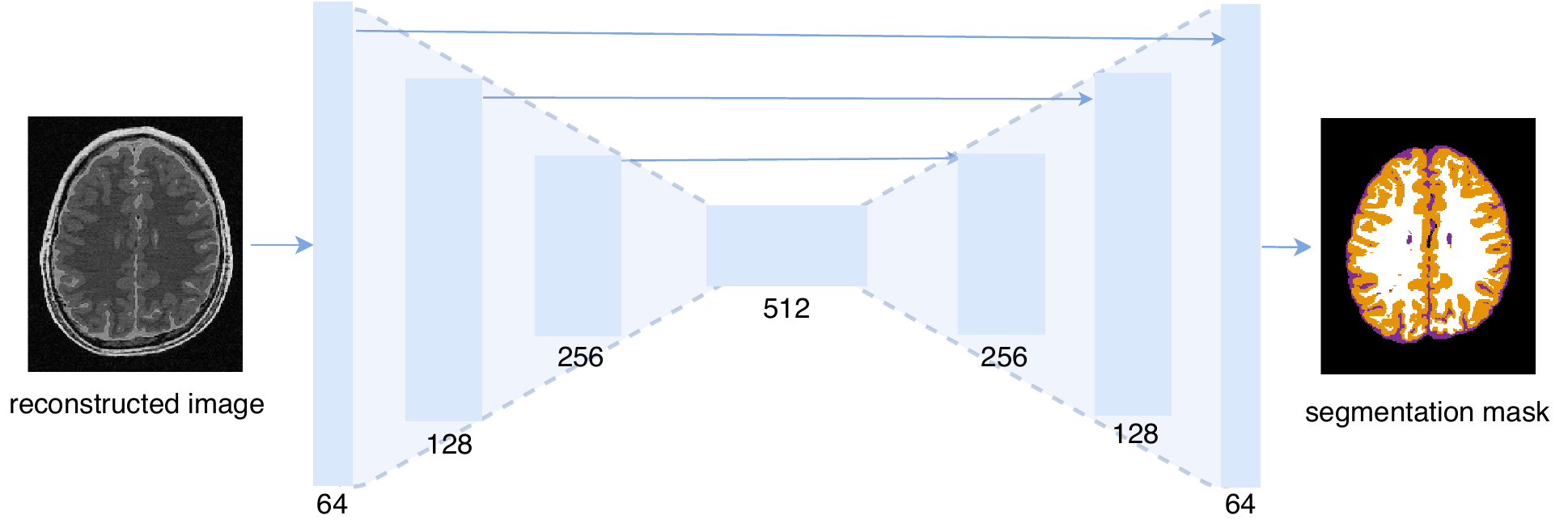} \\
    (c) Segmentation network (UNet)
    \caption{Achitectures of two regularization blocks and one segmentation network utilized in the paper.}
    \label{fig:arch}
\end{figure*}

In our SERANet, we implement two types of regularization blocks: cascaded CNN (Type A) \cite{schlemper2017deep} and auto-encoder (Type B) \cite{jointcsliyan2018}, which are two popular choices for image reconstruction using deep learning.
Their architectures are respectively shown in Figure \ref{fig:arch} (a) and \ref{fig:arch} (b).
We also show the architecture of the UNet adopted in the paper for the segmentation module in Figure \ref{fig:arch} (c).

\section{Additional Quantitative Result}
\begin{figure*}[h!]\label{fig:10_result}
    \centering
    \includegraphics[width=\textwidth]{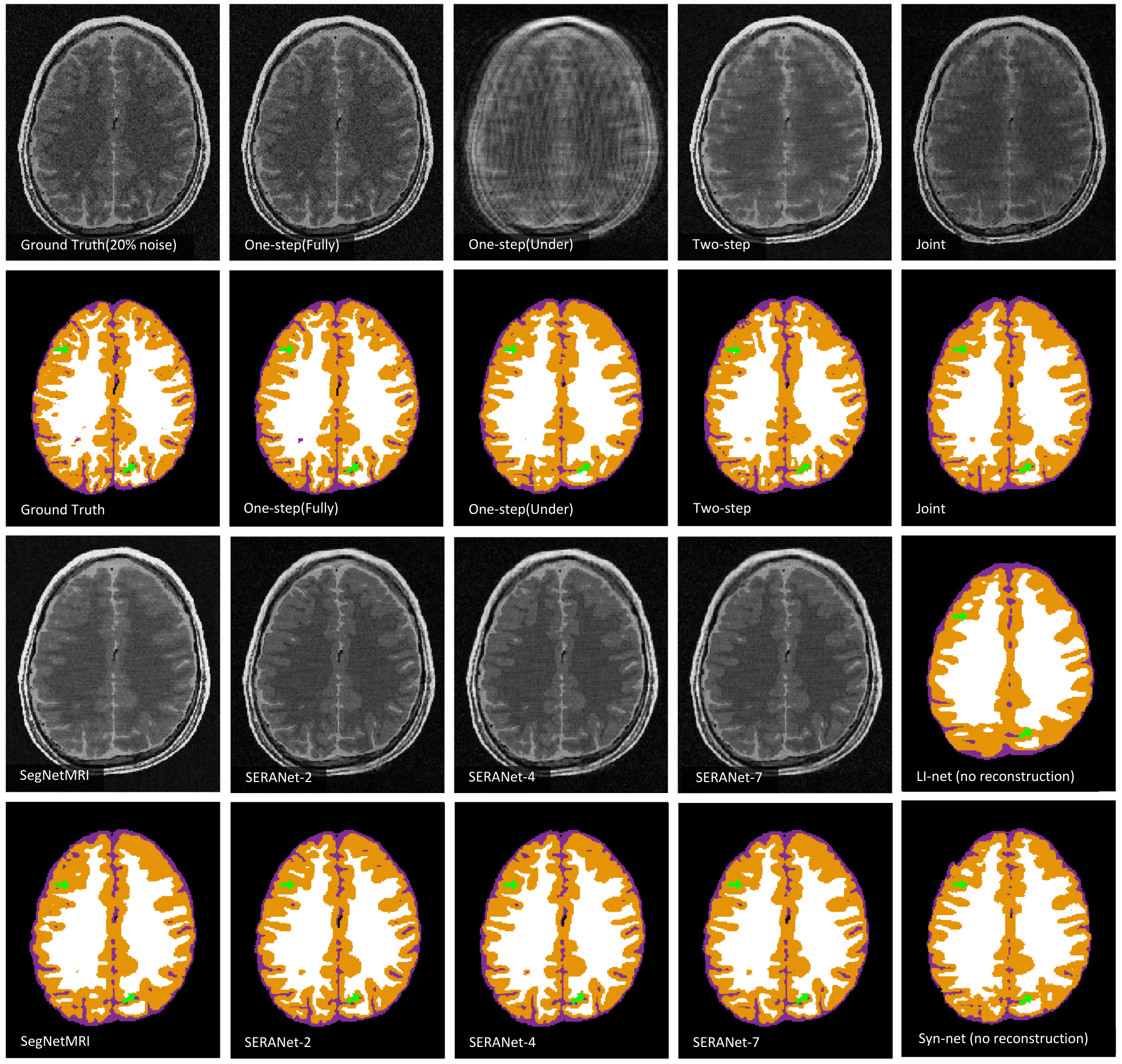}
    \caption{Segmentation performance on input data with $10\%$ white Gaussian noise. Since LI-net and Syn-net bypass the reconstruction, we only show their segmentation results here.}
\end{figure*}

\begin{figure*}[h!]\label{fig:20_result}
    \centering
    \includegraphics[width=\textwidth]{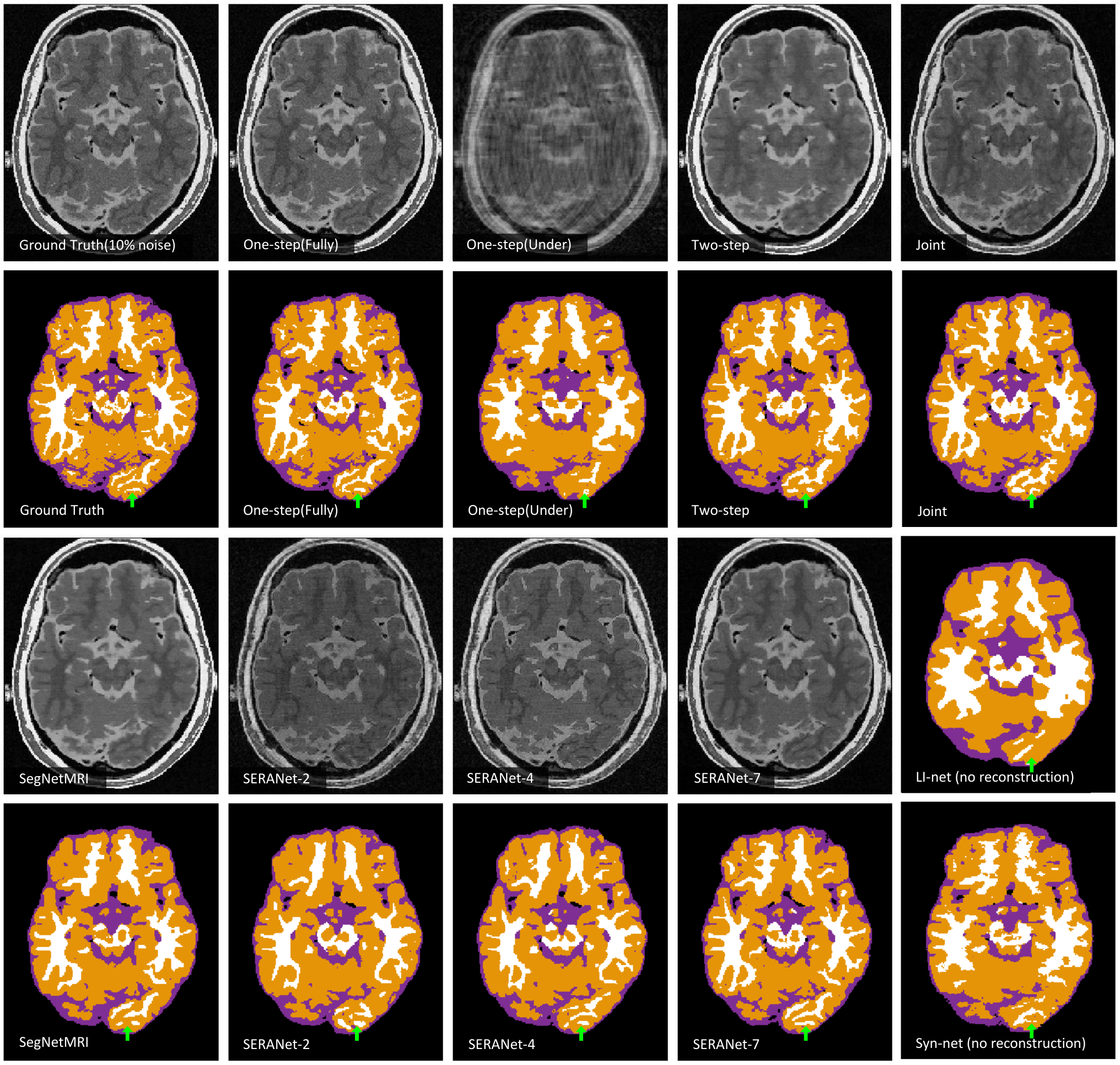}
    \caption{Segmentation performance on input data with $20\%$ white Gaussian noise.
    Since LI-net and Syn-net bypass the reconstruction, we only show their segmentation results here.}
\end{figure*}

In this section, we provide additional quantitative results. 
We demonstrate the comparison results of our SERANet and other approaches on data with $10\%$ (Figure \ref{fig:10_result}) white Gaussian noise and data with $20\%$ noise (Figure \ref{fig:20_result}).
For LI-net and Syn-net, we only show their segmentation results since they bypassed the reconstruction step.
For our SERANet, we present the results of SERANet-2, SERANet-4 and SERANet-7 here.

%
\bibliographystyle{splncs04}
\bibliography{paper1349}
\end{document}